\documentclass[dvipsnames]{article} %
\usepackage{colm2024_conference}

\usepackage{booktabs}
\usepackage{enumitem}
\usepackage{wrapfig}
\usepackage{algorithm}
\usepackage{algpseudocode}
\usepackage{graphicx}
\usepackage[misc]{ifsym}

\usepackage{microtype}
\usepackage{amsmath}
\usepackage{colortbl}
\usepackage[utf8]{inputenc}
\usepackage[T1]{fontenc}
\definecolor{lightgray}{rgb}{0.9,0.9,0.9}
\usepackage{caption}
\usepackage{subcaption}
\usepackage{setspace}
\usepackage{url}
\usepackage{multirow}
\usepackage{tabularx}
\usepackage{blindtext}
\usepackage{pgfplots}
\pgfplotsset{compat=1.18} 
\usepackage{tikz}
\usetikzlibrary{er,positioning,bayesnet}
\usepackage{makecell}
\usepackage{tipa}
\usepackage{siunitx}
\usepackage{nicefrac}
\usepackage{listings}
\usepackage[raster,skins, most]{tcolorbox} %
\usepackage{xltabular}
\usepackage{adjustbox}
\usepackage{xurl}
\usepackage{rotating}
\usepackage[normalem]{ulem}
\usepackage{longtable}

\usepackage{fontawesome}

\useunder{\uline}{\ul}{}

\usepackage{amsmath,amsfonts,bm}

\def\eqref#1{equation~\ref{#1}}

\def\1{\bm{1}}

\DeclareMathAlphabet{\mathsfit}{\encodingdefault}{\sfdefault}{m}{sl}
\SetMathAlphabet{\mathsfit}{bold}{\encodingdefault}{\sfdefault}{bx}{n}

\newcommand*\justify{%
  \fontdimen2\font=0.4em
  \fontdimen3\font=0.2em
  \fontdimen4\font=0.1em
  \fontdimen7\font=0.1em
  \hyphenchar\font=`\-
}

\renewcommand{\texttt}[1]{%
  \begingroup
  \ttfamily
  \begingroup\lccode`~=`/\lowercase{\endgroup\def~}{/\discretionary{}{}{}}%
  \begingroup\lccode`~=`[\lowercase{\endgroup\def~}{[\discretionary{}{}{}}%
  \begingroup\lccode`~=`.\lowercase{\endgroup\def~}{.\discretionary{}{}{}}%
  \catcode`/=\active\catcode`[=\active\catcode`.=\active
  \justify\scantokens{#1\noexpand}%
  \endgroup
}

\usepackage{makecell}
\usetikzlibrary{tikzmark}
\makeatletter
\newcommand*\myfontsize{%
  \@setfontsize\myfontsize{7}{8}%
}
\makeatother

\definecolor{uclablue}{RGB}{159, 195, 224}

\definecolor{uclagold}{RGB}{255, 240, 180}

\definecolor{aliceblue}{RGB}{255, 238, 241}

\definecolor{cadmiumgreen}{rgb}{0.0, 0.42, 0.24}

\definecolor{myred}{rgb}{0.7, 0.3, 0.0}
\definecolor{myblue}{rgb}{0.2, 0.3, 0.6}
\definecolor{babygreen}{rgb}{0.85, 0.97, 0.85}

\definecolor{purple1}{RGB}{126, 107, 196}
\definecolor{purple2}{RGB}{199, 158, 207}
\definecolor{purple3}{RGB}{214, 200, 255}
\definecolor{purple4}{RGB}{254, 240, 255}

\definecolor{deepblue}{RGB}{48, 58, 82}

\newcommand{\symboletongyi}{\raisebox{0pt}{~\includegraphics[scale=0.012]{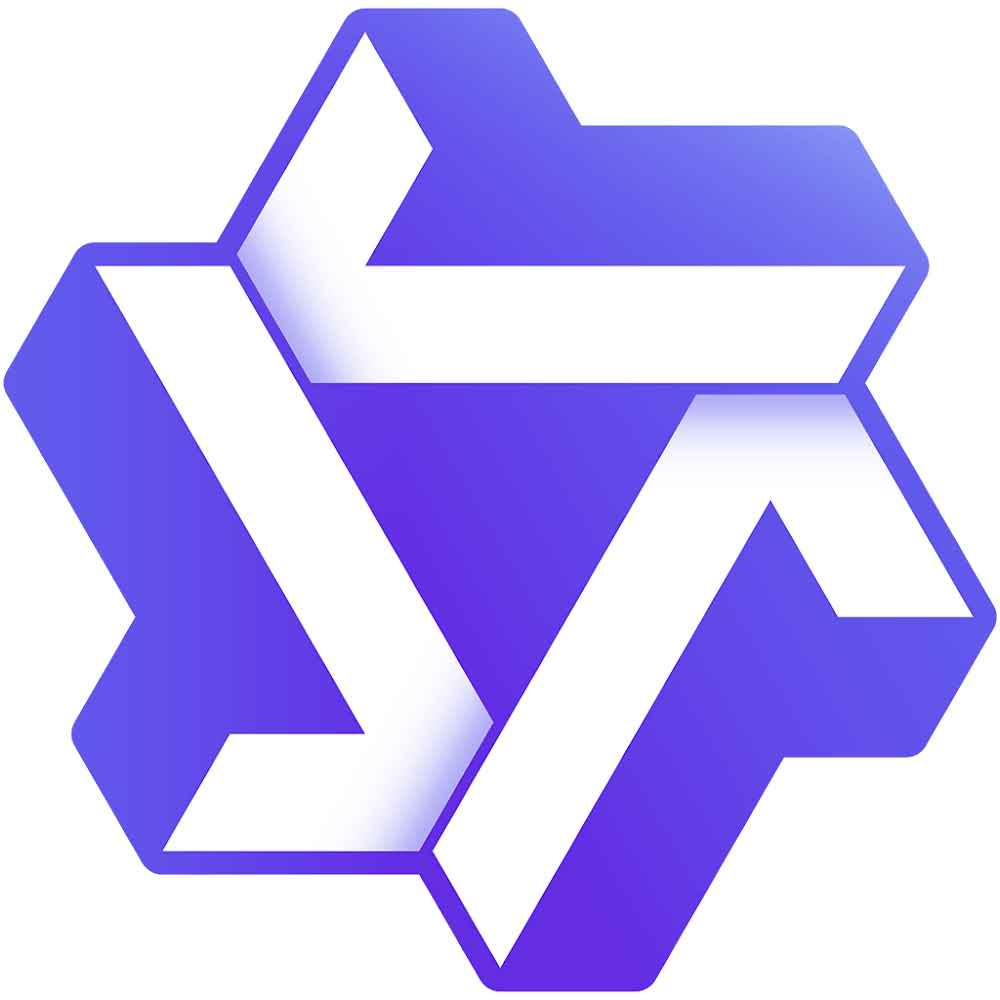}}~}

\definecolor{deepPurple}{HTML}{330066}

\definecolor{uclablue_old}{rgb}{0.15, 0.45, 0.68}
\hypersetup{
    breaklinks,
    citecolor=uclablue_old,
    colorlinks=true,
}

\newtcolorbox{mybox}[2][]
  {colback = black!5!white, colframe = black!75!black, fonttitle = \bfseries,
    colbacktitle = black!100!black, enhanced, before upper={\fontsize{8}{11}\obeyspaces\obeylines\selectfont}, fontupper=\selectfont,
    attach boxed title to top left={yshift=-2.2mm,xshift=4mm},
    title=#2,#1}

\title{%
\raisebox{-2.0em}{
  \parbox[t]{0.35in}{\includegraphics[width=0.6in]{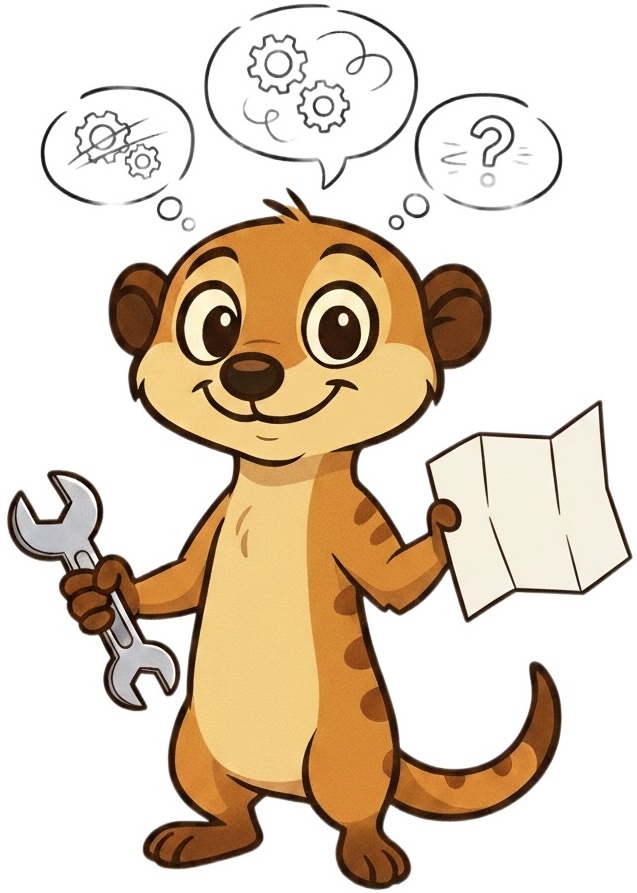}} 
  }
\begin{tabular}[t]{l} 
  \parbox[t]{0.8\textwidth}{\centering 
    AgentFold: Long-Horizon Web Agents with \\Proactive Context Management
  }
\end{tabular}
}

\author{%
\small{Rui Ye$^{*}$$^{(\textrm{\Letter})}$, Zhongwang Zhang\thanks{Equal Core Contributors.}\hspace{0.5mm}, Kuan Li$^{*}$, Huifeng Yin$^{*}$$^{(\textrm{\Letter})}$\\Zhengwei Tao, Yida Zhao, Liangcai Su, Liwen Zhang, Zile Qiao, Xinyu Wang\\Pengjun Xie, Fei Huang, Siheng Chen, Jingren Zhou, Yong Jiang$^{(\textrm{\Letter})}$}%
  \\[1em]               
  {\fontsize{10pt}{11pt}\selectfont          
Tongyi Lab\symboletongyi, Alibaba Group}\\
}

\begin{document}

\maketitle

\begingroup
  \renewcommand\thefootnote{\Letter}  
  \footnotetext{Corresponding Authors. yr991129@sjtu.edu.cn \{yinhuifeng.yhf, yongjiang.jy\}@alibaba-inc.com} 
\endgroup

\begin{abstract}
LLM-based web agents show immense promise for information seeking, yet their effectiveness on long-horizon tasks is hindered by a fundamental trade-off in context management. Prevailing ReAct-based agents suffer from context saturation as they accumulate noisy, raw histories, while methods that fixedly summarize the full history at each step risk the irreversible loss of critical details. Addressing these, we introduce AgentFold, a novel agent paradigm centered on proactive context management, inspired by the human cognitive process of retrospective consolidation. AgentFold treats its context as a dynamic cognitive workspace to be actively sculpted, rather than a passive log to be filled. At each step, it learns to execute a `folding' operation, which manages its historical trajectory at multiple scales: it can perform granular condensations to preserve vital, fine-grained details, or deep consolidations to abstract away entire multi-step sub-tasks. 
The results on prominent benchmarks are striking: with simple supervised fine-tuning (without continual pre-training or RL), our AgentFold-30B-A3B agent achieves \textbf{36.2\%} on BrowseComp and \textbf{47.3\%} on BrowseComp-ZH. Notably, this performance not only surpasses or matches open-source models of a dramatically larger scale, such as the DeepSeek-V3.1-671B-A37B, but also surpasses leading proprietary agents like OpenAI's o4-mini.
\end{abstract}

\begin{figure}[!h]
    \centering
    \includegraphics[width=1.0\linewidth]{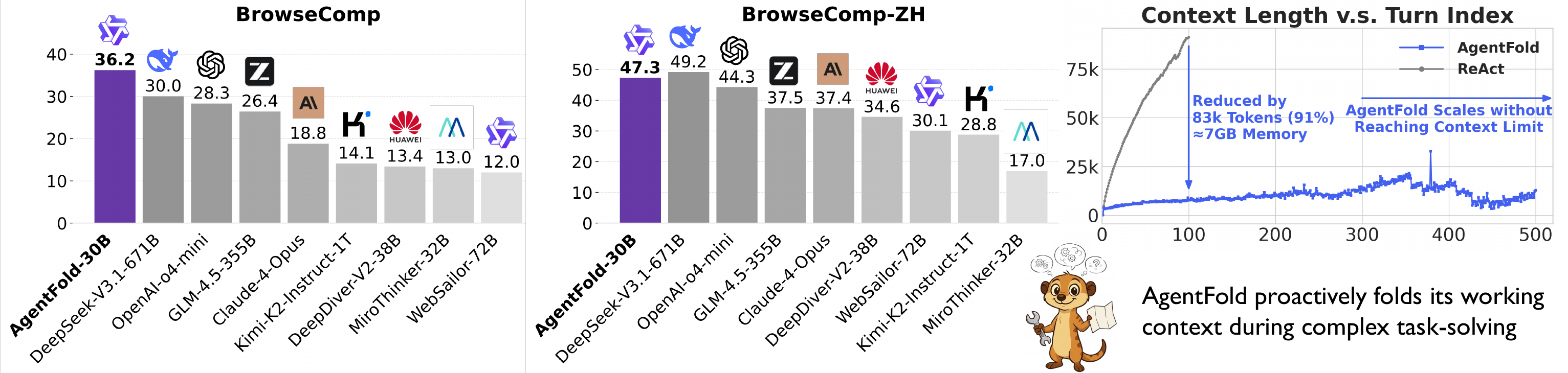}
    \caption{Our AgentFold-30B-A3B agent demonstrates remarkable performance on challenging long-horizon benchmarks, matching or surpassing agents with significantly larger model sizes. This is enabled by its proactive context folding, which maintains a highly concise and focused context that reaches only 7k tokens after 100 turns of interaction and is capable of scaling to 500 turns.}
    \label{fig:teaser}
\end{figure}

\section{Introduction}

The ability to effectively seek and synthesize web information~\citep{marchionini1995information,given2023looking} is foundational to modern progress.
This critical process, however, is fundamentally constrained by inherent human limitations in cognitive capacity and endurance.
The advent of LLM-based web agents marks a paradigm shift, offering systems that transcend these boundaries to tirelessly navigate the digital landscape and dramatically enhancing the efficiency and effectiveness of complex information-seeking tasks~\citep{dr,comanici2025gemini}.

However, a critical challenge for contemporary web agents lies in striking an effective balance between context comprehensiveness and conciseness, a trade-off that significantly impacts their performance, especially on long-horizon tasks~\citep{bc_en,wong2025widesearch}. 
(1) Prevailing ReAct-based agents~\citep{yao2023react,wu2025webdancer,li2025websailor}, which accumulate the entire history of reasoning-action-observation triplets in their context, preserve informational integrity but severely suffer from the overwhelming noise of raw web data, leading to suboptimal actions. 
(2) Conversely, recent approaches~\citep{zhou2025mem1,yu2025memagent,wang2025uitars2} that mechanically summarize the full history at every step maintain a clean context but risk the premature and irreversible loss of crucial details during any single summarization phase. 
These fundamental limitations reveal a critical gap in current methodologies, signaling the necessity for a next-generation agent paradigm with advanced context management.

In this paper, we posit that an ideal agent should manage its internal context like a human's mental scratchpad: a workspace to be actively managed, not passively filled~\citep{miller1956magical}. 
Human problem-solving entails neither the exhaustive retention of all information nor its rigid, step-wise summarization. 
Instead, it is a process of disciplined, retrospective consolidation performed at critical points.
This involves a dynamic `look-back' mechanism: after several actions, irrelevant steps are discarded, intermediate findings are distilled, and key insights are abstracted~\citep{newell1972human}.
This self-correcting act of consolidation is what enables effective and sustained reasoning, a capability we believe is essential for effective long-horizon reasoning and exploration in an agent.

Following this spirit, we introduce \textbf{AgentFold}, an agent architected to proactively and intelligently `fold' segments of context during task execution.
It operates not on a monolithic log, but on a dynamic trajectory composed of \textcolor{blue!70}{\textit{Multi-Scale State Summaries}}—several distilled records of past events—and the \textcolor{red!70}{\textit{Latest Interaction}}, which is the complete record of the most recent action and observation.
At each intermediate step of task-solving trajectory, AgentFold conducts deep reasoning that leads to two concurrent outputs: a folding directive and a tool call.
This folding directive has a dual (two-scale) character:
(1) as a granular condensation, it crystallizes the \textit{Latest Interaction} into a new state summary, appending it to the sequence of \textit{State Summaries};
(2) or as a deep consolidation, it fuses the \textit{Latest Interaction} with a chain of prior summaries, retracting these specific entries and replacing them with a single abstraction at a coarser strategic scale. This is powerful for maintaining logical coherence and conciseness, for instance, by packaging a completed sub-investigation into its final conclusion.
Simultaneously, the resulting observation from the executed tool call then, combined with the action, constitutes the new \textit{Latest Interaction} for the subsequent cycle.
By choosing \textit{what and how much to fold}, AgentFold transcends the brutal trade-off between retaining noisy details and risking catastrophic information loss.
This capability equips AgentFold with a focused and deeply informed reasoning process, essential for conquering long-horizon challenges.

Training AgentFold requires a dataset that does not yet exist: trajectories that demonstrate a sophisticated
interplay of situational action and strategic context curation.
To this end, we develop Fold-Generator, a
specialized LLM-oriented data collection pipeline that can automatically generate trajectories for training.
Recognizing that even the most advanced LLMs cannot reliably produce AgentFold's structured, multi-part responses through prompt engineering along, we leverage a series of rejection sampling mechanism and finally fine-tunes AgentFold based on open-source LLMs.

To validate our folding paradigm, we implement AgentFold by conducting supervised fine-tuning on the Qwen3-30B-A3B model~\citep{yang2025qwen3}. The results on prominent information-seeking benchmarks are striking. Our resulting AgentFold-30B-A3B achieves state-of-the-art performance, scoring \textbf{36.2\%} on BrowseComp~\citep{bc_en}, \textbf{47.3\%} on BrowseComp-ZH~\citep{bc_zh}, \textbf{62.1\%} on WideSearch~\citep{wong2025widesearch}, and \textbf{67.0\%} on general benchmark GAIA~\citep{mialon2023gaia}.
Notably, this performance not only surpasses leading proprietary agents like OpenAI's o4-mini~\citep{o3} but also matches or surpasses open-source models of a dramatically larger scale, such as the GLM-4.5-355B-A32B~\citep{zeng2025glm} and the DeepSeek-V3.1-671B-A37B~\citep{deepseekv3.1}.

\section{Related Works}

\textbf{Web Agents.}
The advent of LLM-based web agents marks a paradigm shift how human seeks information, as these agents could tirelessly and broadly search and synthesize web information.
Pioneering efforts such as OpenAI's deep research~\citep{dr} have demonstrated their promising potential, attracting massive interests from both academia and industry~\citep{zhang2025landscape}.
The majority of contemporary web agents are architected upon the influential ReAct paradigm~\citep{yao2023react}, where an agent iteratively interacts with an environment in a reasoning-action-observation loop.
Examples include WebThinker~\citep{li2025webthinker}, WebDancer~\citep{wu2025webdancer}, WebSailor~\citep{li2025websailor}, WebSailor-V2~\citep{li2025websailorv2}, WebShaper~\citep{tao2025webshaper}, WebExplorer~\citep{liu2025webexplorer} that focus on dataset construction; X-Master~\citep{chai2025scimaster} and BrowseMaster~\citep{pang2025browsemaster} that focus on test-time scaling~\citep{masgpt}.
However, the append-only context inherent to the ReAct paradigm leads to context saturation on long-horizon tasks, impairing reasoning as critical signals become buried in noise. Our work addresses this vulnerability by empowering AgentFold to proactively sculpt its cognitive workspace, ensuring the context remains focused and efficient.

\textbf{Context Management.}
Context management, or context engineering, is an emerging research topic aiming to provide LLM agents with an appropriate and effective context~\citep{context_survey,qiao2025webresearcher}.
A significant line of research focuses on \textit{External Context Augmentation}, which injects relevant knowledge from sources outside the current task trajectory—such as user profiles or past conversations—to provide a richer, more personalized context~\citep{li2025memos,chhikara2025mem0,xu2025amem,memory3}.
Our work, in contrast, pursues \textit{Intra-Task Context Curation}, which focuses on managing the context generated within the task itself to maintain relevance and efficiency over long horizons.
Along this line, MEM1~\citep{zhou2025mem1} and MemAgent~\citep{yu2025memagent} are two recent attempts that compress the full history at each step.
However, these methods employ a rigid, step-wise summarization policy and have been primarily evaluated on simpler, retrieval-focused tasks like HotpotQA~\citep{yang2018hotpotqa}.
Unlike these methods, AgentFold introduces a flexible look-back mechanism that avoids rigid, step-wise compression by retrospectively evaluating and selectively folding multi-step interactions at different scales, a capability crucial for complex, long-horizon tasks~\citep{bc_en,bc_zh}.

\section{AgentFold: Web Agent with Proactive Context Folding}

\subsection{Overview}

AgentFold is a novel web agent designed to tackle complex, long-horizon tasks by emulating a key aspect of human cognition: proactive and structured context/memory management.
At its core, AgentFold makes two primary designs:
first, it defines the agent's context not as a monolithic log, but as a dynamic cognitive workspace.
Second, it empowers the agent to proactively operate upon and sculpt this workspace as an intrinsic part of its reasoning process.

AgentFold's workspace (i.e., context) is explicitly partitioned into the invariant user question, the curated \textcolor{blue!70}{Multi-Scale State Summaries} representing long-term memory, and the high-fidelity \textcolor{red!70}{Latest Interaction} serving as the immediate working memory.
Based on this workspace, the agent's operational process unfolds iteratively. 
In a typical step, its reasoning yields a multi-part response comprising a folding directive to manage historical state summaries, an explanation of its thought process, and the next action.
The folding directive is immediately applied to update the \textcolor{blue!70}{Multi-Scale State Summaries} for future steps, while the explanation, the executed action and its resulting observation form the new \textcolor{red!70}{Latest Interaction} for the subsequent cycle.
This process repeats until the agent determines it has gathered sufficient information to provide an accurate final answer, with the initial step being a special case that omits the folding directive due to the absence of prior history.

This operational cycle establishes a powerful \texttt{perceive -> reason -> fold -> act} loop, where context curation is an explicit, learned step rather than a passive byproduct. 
By synergizing a well-defined cognitive workspace with the agent's autonomy to manipulate it, AgentFold directly resolves the critical trade-off between retaining granular details and preventing context inflation, enabling a more focused and efficient reasoning process for complex, long-horizon challenges.

\begin{figure}[t]
    \centering
    \includegraphics[width=1.0\linewidth]{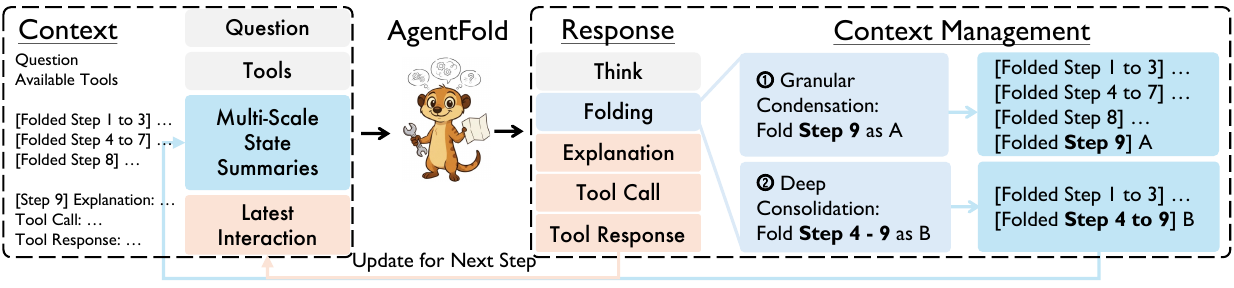}
    \caption{Overview of AgentFold at an intermediate step. The two key parts in AgentFold' context are: \textcolor{blue!70}{Multi-Scale State Summaries} (several folded blocks recording previous information) and \textcolor{red!70}{Latest Interaction} (a full record of the latest step). AgentFold responds with four blocks: thinking, folding, explanation, and tool call (which leads to an appended tool response). The folding directive has two operation modes: granular condensation that folds one single step with useful information reserved and deep consolidation that folds several steps with a coarse summary especially when these steps complete a sub-task and the intermediate details are not critical for further task-solving.}
    \label{fig:agentfold_overview}
\end{figure}

\subsection{AgentFold's Context: Multi-Scale State Summaries, Latest Interaction}

The performance of a web agent is critically dependent on the quality and structure of the context it receives.
To this end, we design AgentFold's context as a dynamic cognitive workspace partitioned into four distinct components (i.e., user question, available tools, multi-scale state summaries, and latest interaction) to enable both strategic long-range planning and precise situational action. 

(1) The question serves as an anchor, constantly reminding the agent of its ultimate objective.
(2) The list of available tools defines the agent's capacity for action within its environment.
This component provides a detailed schema for each tool—including its name, description, and required parameters—outlining the agent's entire suite of executable operations.
(3) The \textcolor{blue!70}{Multi-Scale State Summaries} function as the agent's curated long-term memory.
This component preserves the sequential, logical flow of the trajectory, with past steps recorded at different scales based on their perceived utility for future actions. 
This multi-scale structure allows critical findings to be retained as distinct, fine-grained summaries, while less critical intermediate steps can be consolidated into coarser, more abstract blocks.
Consequently, it retains a coherent historical narrative while minimizing informational noise. 
(4) The \textcolor{red!70}{Latest Interaction} acts as a high-fidelity working memory.
It provides the complete, complete record of the most recent transaction—including the agent's brief thinking (explanation), the executed tool call, and the resulting observation.
This full transparency into the immediate past is crucial for providing the situational awareness needed to make a sound decision, which involves both how to selectively fold this new information and what action to generate next.
The entire architecture mirrors how humans leverage a stable goal, consolidated knowledge, and a volatile working memory.

Specifically, the context $C_t$ provided to the agent at step $t$ is a triplet:
\begin{equation}
    C_t = (Q, T, S_{t-2}, I_{t-1})
\end{equation}
where $Q$ and $T$ are the invariant user question and tools, respectively.
$S_{t-2}$ represents the \textcolor{blue!70}{Multi-Scale State Summaries}, a dynamically updated sequence of condensed summaries from previous steps. Formally, we represent $S_{t-2}$ as an ordered sequence of summary blocks:
\begin{equation}
    S_t = (s_{x_1, y_1}, s_{x_2, y_2}, \ldots, s_{x_m, y_m})
\end{equation}
where each $s_{x,y}$ is a textual summary of the contiguous block of steps from $x$ to $y$. The step ranges partition the entire history up to the previous step, such that $x_1=1$, $y_m=t-2$, and $x_{i+1} = y_i + 1$ for all $i$. This notation explicitly captures the \textbf{multi-scale} property: a summary of a single, independent step is denoted as $s_{x,x}$ (where $y=x$), whereas a summary representing the consolidation of a multi-step process (e.g., verifying a condition over several steps) is denoted as $s_{x,y}$ (where $y>x$). The third component, $I_{t-1}$, is the \textcolor{red!70}{Latest Interaction}, a verbose, complete record of the previous step's full transaction. It is formed by concatenating the explanation, action, and observation from step $t-1$: $I_{t-1} = (e_{t-1}, a_{t-1}, o_{t-1})$. For the initial step ($t=1$), the context is initialized as $C_1 = (Q, T, \emptyset, \emptyset)$, containing only the user's question; while for step 2, the context is set as $C_2 = (Q, T, \emptyset, I_1)$, containing the user question and latest interaction.

This structured context design offers the best of both worlds.
The \textcolor{red!70}{Latest Interaction} provides the raw, granular detail necessary for the agent to make informed, short-term decisions without any loss of information.
Simultaneously, the \textcolor{blue!70}{Multi-Scale State Summaries} offer a noise-free, abstracted overview of the mission so far, preventing the agent from getting lost in irrelevant details and enabling coherent long-term reasoning.
This structure directly mitigates the trade-off between context comprehensiveness and conciseness that hinges contemporary web agents.

\subsection{AgentFold's Response: Thinking, Folding, Explanation, Action}

Complementing its structured cognitive workspace, AgentFold's response is not a monolithic command but a multi-faceted output that reflects its dual role as both a situational problem-solver and a strategic context manager.
At each step, the agent generates a single, coherent block of text that is parsed into three components, each designed to operate on its context: a directive to \textit{fold} its long-term memory, an \textit{explanation} to articulate its motivation behind the following action, and an \textit{action} to propel the task forward.
This design integrates context management as a core, learnable component of the agent's reasoning process, rather than treating it as a passive byproduct.

Specifically, at each step $t$, AgentFold generates a response $R_t$ based on the context $C_t$ and model $\theta$. This response is a single, coherent block of text designed to be parsed into a quadruplet:
\begin{equation}
    R_t = \text{AgentFold}(C_t; \theta) \rightarrow (th_t, f_t, e_t, a_t)
\end{equation}
Here, $th_t$ is the thinking process, a detailed chain-of-thought monologue where the agent analyzes its context ($C_t$) and weighs options for both context folding and the subsequent action. From this internal deliberation, the other three structured components are derived. 
(1) The \textbf{folding directive} ($f_t$) is the agent's explicit command for sculpting its \textcolor{blue!70}{Multi-Scale State Summaries} $S_{t-2}$. It takes the form of a JSON object: 
$f_t = \mathtt{\{``range":[}$k$\mathtt{, } $t-1$\mathtt{],``summary":``}\sigma_t\mathtt{"\}}$, where $k$ is the starting ID for folding and $\sigma_t$ is the replacement summary text that can be proactively determined by AgentFold itself.
This single format supports two modes of context management that operate at different scales:
\begin{itemize}[leftmargin=*]
    \item[\textbullet] \textbf{Granular Condensation} ($k=t-1$): This operation folds only the \textit{Latest Interaction} into a new, fine-grained summary. It is used for incremental steps, preserving the highest resolution of the historical trajectory by converting a single verbose record into a compact summary block (e.g., `[Compressed Step 5] Found a new candidate XYZ that needs further exploration.').
    \item[\textbullet] \textbf{Deep Consolidation} ($k < t-1$): This operation performs a change of scale by fusing the \textit{Latest Interaction} with a chain of prior summaries into a single, coarse-grained summary. This is powerful for abstracting away noisy, intermediate steps once a sub-task is complete. For instance, an agent might spend multiple steps verifying a single fact, navigating through irrelevant websites or encountering failed tool calls. Deep consolidation allows the agent to retract this entire verbose sequence and replace it with a single, conclusive summary (e.g., `[Compressed Step 5 to 9] Confirmed that XYZ does not fit all criteria after checking several sources]').
\end{itemize}
This directive transforms \textcolor{blue!70}{Multi-Scale State Summaries} $S_{t-2}$ into $S_{t-1}$ by retracting all summary blocks whose steps fall within the range $[k, t-1]$ and replacing them with a single new summary block, $s_{k, t-1}=\sigma_t$.
(2) The explanation ($e_t$) is a concise summary of the key insights from the thinking process, articulating the motivation for the chosen action. 
(3) Finally, $a_t$ is the agent's chosen external action, which is either a tool call with specified tool name and tool arguments, or the final answer if the agent deems that no further interaction is required.
When another tool call is invoked, the tool will be executed to obtain the observation $o_t$ from the environments.
Finally, the question $Q$, tools $T$, the new \textcolor{blue!70}{Multi-Scale State Summaries} $S_{t-1}$ and \textcolor{red!70}{Latest Interaction} $I_{t}=(e_t, a_t, o_t)$ constitute the AgentFold's context $C_{t+1}=(Q,T,S_{t-1},I_t)$ for the next step $t+1$.

This structured response architecture engenders a powerful cognitive \textbf{symbiosis} between the agent's two core deliberations: planning the next action and curating its own context.
(1) The explicit requirement to formulate a folding directive compels the agent to critically evaluate its trajectory and distill the most salient information from its historical context.
This act of reflection inherently sharpens its understanding of the current state, leading to a more informed and effective subsequent action. (2) Conversely, the process of planning a new action necessitates a purposeful interrogation of its recent history to identify pivotal clues. 
This very process of determining what is immediately relevant provides a perfect, real-time signal for what is worth preserving in a folded summary. 
This tight coupling of acting and reflecting ensures that AgentFold's behavior is both purposeful and efficient, creating a self-regulating loop that simultaneously enhances the quality of its actions and the coherence of its context memory.

\subsection{AgentFold's Training: Data Trajectory Collection}

Training AgentFold requires a dataset that does not yet exist: trajectories that demonstrate a sophisticated interplay of situational action and strategic context curation.
To this end, we develop \textit{Fold-Generator}, a specialized data collection pipeline built upon powerful open-source Large Language Models (LLMs) to generate the necessary trajectory training data.
To ensure a fair and direct comparison with prior work, we utilize the same question set as the recent WebSailor work~\citep{li2025websailorv2}.
We find that even the most advanced LLMs cannot reliably produce AgentFold's accurate, structured, multi-part responses through prompt engineering alone.
To relieve the effects of this, we leverage a rejection sampling mechanism, discarding any generated step that fails to strictly adhere required formats, or any trajectory that contains too many environmental errors.
This ensures every data point in our collection is a clear example of the desired reasoning process.

Specifically, the output of the Fold-Generator is a collection of high-quality interaction pairs, $\{ (C_t, R_t^*) \}_N$, where each $C_t$ is the structured context, $R_t^*$ is the validated, gold-standard response, and $N$ is the total number of interaction steps across all questions. 
This curated dataset is then used for conducting conventional Supervised Fine-Tuning (SFT) on open-source LLMs.
The training objective is to distill the complex, multi-step, validated reasoning of our pipeline into a single, efficient forward pass, thereby teaching the model to produce the entire structured output intrinsically.

This training methodology is not merely an implementation choice but a necessity that yields critical advantages.
Primarily, it transforms the agent's ability to `fold` from a fragile, prompt-dependent instruction into a robust, internalized skill. 
Furthermore, the SFT process effectively distills the computationally intensive `generate-and-filter` strategy into the weights of the final AgentFold model.
This results in a specialized agent that is not only highly capable but also significantly more efficient at inference time than the general-purpose models used for data collection.
Finally, by building this entire pipeline on open-source models, we maintain full transparency and control over the data and training process, enabling detailed inspection and future iteration.

\subsection{Discussions}

AgentFold's design offers a novel approach to context management, resolving the trade-off between the append-only history of ReAct, which leads to context saturation, and uniform full-history summarization, which risks irreversible information loss.
The primary advantage lies in the agent's ability to adapt its folding strategy. It can employ \textbf{Granular Condensation} to preserve a potentially vital, fine-grained detail, protecting it from the indiscriminate compression of a full-history summarizer. Conversely, it can use \textbf{Deep Consolidation} to prune an entire concluded sub-investigation, combating the noise accumulation found in ReAct. Crucially, the ability to delay consolidation until a sub-task's outcome is clear allows for more informed and less short-sighted curation decisions.

This flexibility is critical for maintaining long-term informational integrity. To illustrate, if we assume a modest 1\% chance of a key detail being lost each time the full history is re-summarized, \textbf{the probability of a finding from step 1 surviving until step 100 reduces to just} $\mathbf{\approx 36.6\% (0.99^{100}})$.
This risk is exacerbated in extremely long-horizon tasks; after 500 steps, for instance, the survival probability for the same detail collapses to only \textbf{0.66\%} ($\mathbf{0.99^{500}}$).
AgentFold's \textit{Granular Condensation} directly mitigates this compounding risk by preserving the detail in a distinct block, exempting it from unnecessary reprocessing.
In parallel, ReAct's append-only policy faces the deterministic certainty of context saturation, where after 100 steps the context is burdened by the full verbosity of every past interaction. AgentFold's \textit{Deep Consolidation} addresses this by surgically pruning such irrelevant traces, ensuring the context remains both focused and computationally manageable.

This represents a conceptual leap from agents with static, predefined context policies to those as \textbf{self-aware knowledge managers}. By integrating context curation as a learnable, core action, AgentFold learns sophisticated, task-specific strategies for what to remember, what to abstract, and what to discard. This ability to actively shape its own informational workspace is the key to its enhanced robustness and efficiency, enabling it to dynamically balance the need for granular detail with a coherent long-term plan on complex, long-horizon challenges.

\section{Experiments}

\textbf{Implementation.}
We train our AgentFold based on open-source LLM Qwen3-30B-A3B-Instruct-2507~\citep{yang2025qwen3} with 30B parameters in total and 3B activated during prediction.
We set the max tool call number as 100, any trajectory beyond this number will be forcibly terminated.

\textbf{Benchmarks.}
We consider 3 information-seeking benchmarks including BrowseComp~\citep{bc_en}, BrowseComp-ZH~\citep{bc_zh}, and WideSearch-en (the most detailed metric: Item-F1)~\citep{wong2025widesearch}; and 1 general benchmark: GAIA (text-only subset)~\citep{mialon2023gaia}.
Note that BrowseComp and BrowseComp-ZH mainly evaluates the agent's capability in locating hard-to-find information; WideSearch emphasizes on capability of broad search; and GAIA is for evaluating general capabilities of AI agents.
For benchmarks with less than 200 samples, we report the averaged results on 3 trials.

\textbf{Baselines.}
We comprehensively compare our AgentFold-30B-A3B with representative open-source agents including WebThinker~\citep{li2025webthinker}, WebDancer~\citep{wu2025webdancer}, WebSailor~\citep{li2025websailor}, ASearcher~\citep{gao2025beyond}, MiroThinker~\citep{2025mirothinker}, WebExplorer~\citep{liu2025webexplorer}, DeepDive~\citep{deepdiver}, DeepDiver-V2~\citep{deepdiver2}, Kimi-K2-Instruct~\citep{kimi-k2}, GLM-4.5~\cite{zeng2025glm}, and DeepSeek-V3.1~\citep{deepseekv3.1}.
We also report performances of several proprietary agents for reference, including Claude-4-Sonnet/Opus~\citep{claude4}, OpenAI-o4-mini/o3~\citep{o3} and OpenAI Deep Research~\citep{dr}.
Some results are taken from corresponding papers or leaderboards.

\subsection{Results and Analysis}

\textbf{Main results}, presented in Table~\ref{tab:main}, demonstrate that AgentFold-30B-A3B establishes a new state of the art for open-source agents and is highly competitive with leading proprietary systems. Notably, it solidifies its dominance in the open-source landscape by outperforming models up to 20 times its size, scoring \textbf{36.2\%} on BrowseComp against the 671B DeepSeek-V3.1's 30.0\%. Furthermore, AgentFold proves its capability at the highest level by achieving the best overall score of \textbf{62.1\%} on WideSearch, surpassing all proprietary agents including OpenAI-o3 and Claude-4-Sonnet. These results underscore the profound impact of our architectural innovations, showcasing how effective context management can bridge the performance gap with dramatically larger models.

\begin{table}[t]
    \caption{Main results. AgentFold-30B-A3B achieves remarkable performance, surpassing open-source agents with much larger model size such as DeepSeek-V3.1-671B-A37B and matching proprietary agents such as OpenAI-o4-mini, indicating the potential of this new paradigm.} 
    
    \centering
    \resizebox{\textwidth}{!}{\begin{tabular}{l|cccc}
    \toprule
    \textbf{Agent} & \textbf{BrowseComp} & \textbf{BrowseComp-ZH} & \textbf{WideSearch} & \textbf{GAIA} \\
    
    \midrule
    \rowcolor{blue!10}\multicolumn{5}{c}{\emph{\textbf{Proprietary Agents}}} \\
    \midrule

    Claude-4-Sonnet & 14.7 & 22.5 & \textbf{62.0} & 68.3 \\
    Claude-4-Opus~\citep{claude4} & 18.8 & 37.4 & - & - \\
    OpenAI-o4-mini~\citep{o3} & 28.3 & 44.3 & - & -  \\
    OpenAI-o3~\citep{o3} & 49.7 & \textbf{58.1} & 60.0 & \textbf{70.5}  \\
    OpenAI Deep Research~\citep{dr} & \textbf{51.5} & 42.9 & - & 67.4 \\
    
    \midrule
    \rowcolor{blue!10}\multicolumn{5}{c}{\emph{\textbf{Open-Source Agents}}} \\
    \midrule

    WebThinker-32B~\cite{li2025webthinker} & 2.8 & 7.3 & - & 48.5 \\ 
    WebDancer-32B~\citep{wu2025webdancer} & 3.8 & 18.0& - & 51.5 \\
    WebSailor-32B~\citep{li2025websailor} & 10.5 & 25.5 & - & 53.2 \\
    WebSailor-72B~\citep{li2025websailor} & 12.0 & 30.1 & - & 55.4 \\
    ASearcher-Web-32B~\citep{gao2025beyond} & 5.2 & 15.6 & - & 52.8  \\
    MiroThinker-32B-DPO-v0.2~\citep{2025mirothinker} & 13.0 & 17.0 & - & 64.1  \\
    WebExplorer-8B~\citep{liu2025webexplorer} & 15.7 & 32.0 & - & 50.0  \\
    DeepDive-32B~\citep{lu2025deepdive} & 14.8 & 25.6 & - & -  \\
    DeepDiver-V2-38B~\citep{deepdiver2} & 13.4 & 34.6 & - & -  \\
    Kimi-K2-Instruct-1T~\citep{kimi-k2} & 14.1 & 28.8 & 59.9 & 57.3  \\
    GLM-4.5-355B-A32B~\citep{zeng2025glm} & 26.4 & 37.5 & - & 66.0  \\
    DeepSeek-V3.1-671B-A37B~\citep{deepseekv3.1} & 30.0 & \textbf{49.2} & - & 63.1 \\
    
    \midrule
    \textbf{AgentFold-30B-A3B (Ours)} & \textbf{36.2} & 47.3 & \textbf{62.1} & \textbf{67.0} \\
    \bottomrule
    \end{tabular}}
    \label{tab:main}
\end{table}

\begin{figure}[t]
    \centering 

    \begin{subfigure}{0.48\linewidth}
        \centering
        \includegraphics[width=\linewidth]{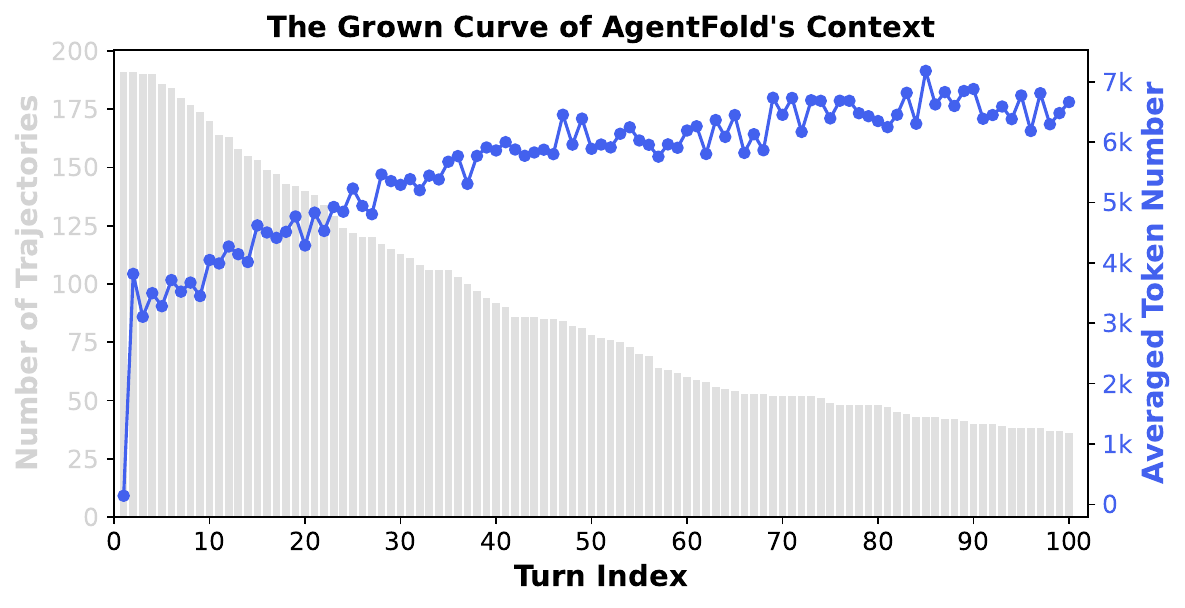}
        \caption{Growth curve of AgentFold's context}
        \label{fig:context_growth}
    \end{subfigure}
    \hfill 
    \begin{subfigure}{0.48\linewidth}
        \centering
        \includegraphics[width=\linewidth]{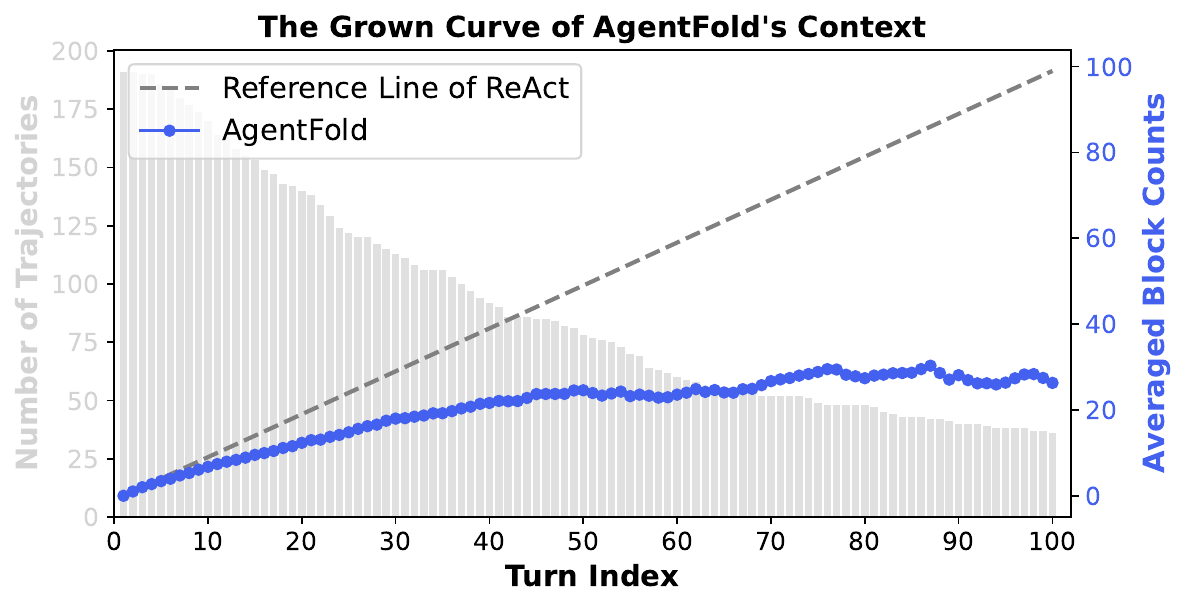}
        \caption{Number of blocks in AgentFold's context}
        \label{fig:block_growth}
    \end{subfigure}
    \caption{Analysis of AgentFold's context on trajectories sampled from BrowseComp. (a) AgentFold's context length grows at a remarkably slow, sub-linear rate, less than doubling from approximately 3.5k to 7k over 100 turns. As our model's max context is 128k, this indicates a promising potential for AgentFold for tackling complex and long-horizon tasks. (b) Our Deep consolidation operation in AgentFold merges multiple past steps into a single summary, thereby maintaining a significantly more structural and concise context compared to the popular ReAct.}
    \label{fig:context}
\end{figure}

\textbf{Dynamics of AgentFold's context: token count.}
To empirically validate AgentFold's context management, we analyze 200 trajectories from the BrowseComp benchmark (Figure~\ref{fig:context_growth}). We plot the number of surviving trajectories at each turn ($|\mathcal{T}_t|$, grey bars) alongside the average context token count ($A_t$, blue curve) for those same trajectories. Specifically, $A_t$ is formally defined as:
$A_t = \frac{1}{|\mathcal{T}_t|} \sum_{j \in \mathcal{T}_t} \text{TokenCount}(C_{j,t})$
where $\mathcal{T}_t$ is the set of surviving trajectories that are consisted of more than $t$ turns, and $C_{j,t}$ is the context of trajectory $j$ at turn $t$.

The figure reveals that AgentFold maintains an \underline{exceptionally concise context}. The average token count grows at a remarkably slow, sub-linear rate, less than doubling from approximately 3.5k to 7k over 100 turns, proving the efficacy of the `fold' operation in preventing context inflation.

When observing the survival curve, we notice that over 20\% of tasks being forcibly terminated at our experimental limit of 100 turns, which are typically marked as failure.
Crucially, at this termination point, the agent's context is only $\approx$7k tokens—a minor fraction of the underlying model's 128k capacity.
This vast remaining capacity points to two promising conclusions.
(1) First, it suggests that simply scaling the number of allowed interactions could unlock higher success rates. 
(2) Second, and more broadly, it demonstrates AgentFold's significant \underline{potential for tackling extremely complex} and long-horizon problems.
We provide a conceptual verification in the following Figure~\ref{fig:scaling_too_call} but defer detailed explorations to future work due to time constraints.

\textbf{Dynamics of AgentFold's context: block count.}
To analyze the structural complexity of the context, we measure the number of discrete `blocks' in the agent's workspace at each turn. A block is defined as any single entry in the \textit{Multi-Scale State Summaries} (e.g., `[Compressed Step 52 to 67]' is one block) plus the one \textit{Latest Interaction}. For an append-only method like ReAct, this count necessarily increases linearly with each turn (reference line in Figure~\ref{fig:block_growth}). The analysis of AgentFold's block count reveals two key conclusions.
(1) Sub-linear growth and structural simplicity. In stark contrast to the linear explosion of ReAct, AgentFold's block count grows at a slow, sub-linear rate. This efficiency is driven by the \underline{Deep Consolidation} operation, which merges multiple past steps into a single summary, thereby maintaining a structurally simple and cognitively manageable context.
(2) Compounding efficiency over time. The growing divergence between the two curves highlights the compounding advantage of proactive curation. While ReAct's append-only policy leads to runaway structural complexity over long horizons, AgentFold's consolidation ensures the context remains controlled, so its efficiency \underline{advantage over ReAct grows larger on longer tasks}.

\textbf{Context comparison between AgentFold and ReAct.}
To provide a more direct and intuitive comparison, we plot the average context length (in tokens) of AgentFold against a standard ReAct baseline across the same set of trajectories. As illustrated on the right of Figure~\ref{fig:teaser}, the contrast is stark. The ReAct agent's context exhibits an uncontrolled, near-linear growth, accumulating a massive token count as the task progresses. In contrast, AgentFold's context size remains remarkably flat and controlled due to its proactive folding mechanism.

By the 100th turn, this architectural difference results in a dramatic quantitative advantage: AgentFold's context is, on average, over \textbf{84k tokens (92\%) smaller} than ReAct's. This token reduction also has profound implications for computational resource requirements, translating to an estimated memory saving of nearly \textbf{7GB per inference instance} at this trajectory length. This analysis demonstrates not only the conceptual benefits of our approach but also its immense practical value in making long-horizon agents more efficient, scalable, and cost-effective.

\begin{wrapfigure}{r}{0.5\textwidth}
  \vspace{-0.1cm}
  \includegraphics[width=0.5\textwidth]{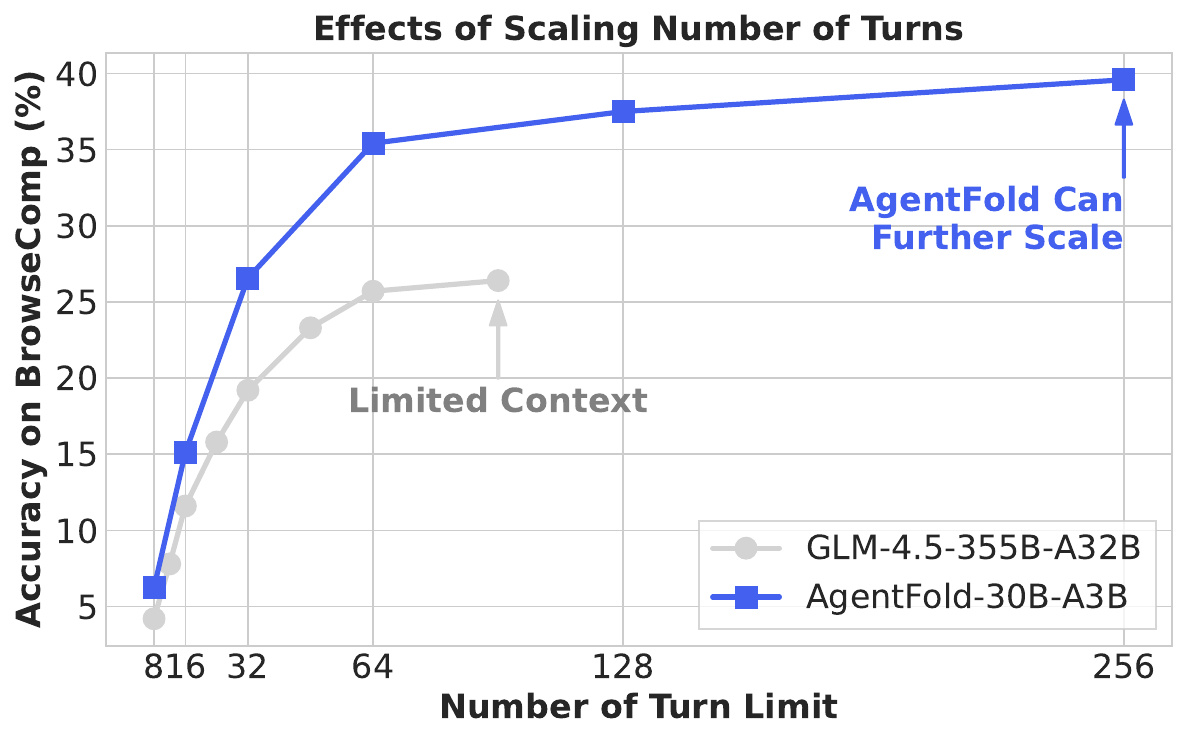}
  \vspace{-5mm}
   \caption{Scaling properties of interaction turns (tool calls). This demonstrates the profound potential of AgentFold to tirelessly and robustly work for hundreds of steps for humans.}
   \vspace{-5mm}
  \label{fig:scaling_too_call}
\end{wrapfigure}
\textbf{Scaling Properties of Interaction Turns.}
Building on our finding of its compact context in Figure~\ref{fig:context_growth}, we test AgentFold's performance when scaling the number of interaction turns, a primary bottleneck for conventional agents. As shown in Figure~\ref{fig:scaling_too_call}, we evaluate on BrowseComp with a turn limit up to 256, comparing our 30B model against a much larger 355B GLM-4.5 baseline. The results show two clear advantages. (1) First, our smaller model consistently outperforms the 355B baseline at all comparable turn limits. (2) Second, the GLM-4.5 agent's performance saturates and fails beyond 64 turns as its append-only context fills, while AgentFold's accuracy continues to improve steadily up to 256 turns, \underline{showing promising scaling property}.

To further probe these limits, we conduct an extended experiment, increasing the maximum number of turns to 500, with the context length dynamics reported in Figure~\ref{fig:teaser}.
The results reveal that the context mostly remains below 20k tokens and, notably, does not grow monotonically.
This behavior is a direct result of AgentFold's ability to recognize and recover from dead ends. When a lengthy line of inquiry proves unsuccessful, the agent can perform a deep consolidation of the entire failed sub-trajectory. This act of abstracting away a long, irrelevant history while pivoting to a new strategy often resets the context to a more compact state, showcasing a sophisticated, self-correcting form of context management.

This experiment confirms that AgentFold's proactive context management is the key to unlocking long-horizon task-solving.
It demonstrates the profound potential for agents to engage in truly extended interactions—potentially lasting for hundreds of steps—to perform the kind of broad and deep web exploration required for complex research and analysis tasks that remain far beyond the reach of conventional agent architectures.

\begin{figure}[t]
    \centering
    \includegraphics[width=1.0\linewidth]{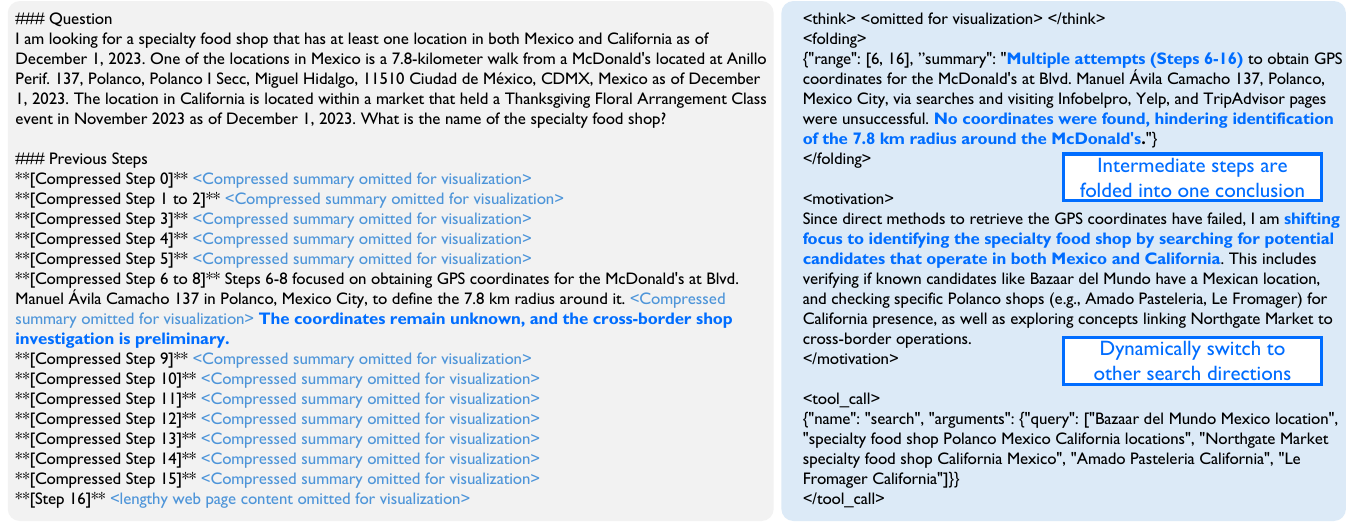}
    \caption{Case study for illustration of AgentFold. See detailed content in Table~\ref{tab:case1_traj}, Figure~\ref{fig:case1_context} and \ref{fig:case1_response}. After a series of failure attempts happened (steps 6 to 16), AgentFold notices that this direction might be a dead end, folds these intermediate steps into one conclusion, plans to switch to other search directions, and decides the new search queries.}
    \label{fig:case_study_main}
\end{figure}

\textbf{Case study.}
To directly illustrate AgentFold's operational intelligence, we provide a case study in Figure~\ref{fig:case_study_main} (with more in Appendix~\ref{app:case_study}).
The figure captures the agent at step 17 of a complex task, where its context already showcases its multi-scale structure, comprising both fine-grained, single-step summaries (e.g., [Compressed Step 5]) and previously consolidated blocks (e.g., [Compressed Step 6 to 8]).
The figure showcases a critical moment of reflection and re-planning.
Recognizing a long and unsuccessful series of attempts (from step 6 to 16) as a dead end, AgentFold executes a decisive, strategic move. First, it performs a \textbf{Deep Consolidation}, folding the entire 11-step failed sequence into a single, conclusive summary.
This operation distills the valuable lesson from the failures—that this approach is unworkable—while pruning the noisy and now-irrelevant procedural details from its context.
Informed by this newly consolidated insight, the agent then dynamically plans (in the motivation block, equals to explanation) to shift towards a new line of investigation, which is immediately reflected in its subsequent tool call.
This example powerfully demonstrates AgentFold's ability to reason about its own trajectory, learn from extended failures, and strategically re-plan by actively curating its cognitive workspace.

\section{Conclusions}

This paper introduces AgentFold, a novel agent paradigm that resolves the fundamental trade-off between context saturation in append-only agents (e.g., ReAct) and irreversible information loss from uniform summarization. We move beyond these static policies by empowering the agent to act as a self-aware knowledge manager, equipped with a proactive `fold` operation to dynamically sculpt its context at multiple scales. This mechanism allows the agent to preserve fine-grained details via \textit{Granular Condensation} while abstracting away irrelevant history with \textit{Deep Consolidation}. Our experiments validate the power of this approach: the AgentFold-30B-A3B model establishes a new state of the art for open-source agents, outperforming models over 20 times its size like DeepSeek-V3.1-671B and proving highly competitive against leading proprietary agents such as OpenAI's o4-mini. Furthermore, its exceptional context efficiency enables truly long-horizon problem-solving by supporting hundreds of interaction steps within a manageable context.

\textbf{What's next.}
In this work, we prioritize demonstrating the potential of the AgentFold paradigm, thus employing a straightforward SFT approach without extensive optimization. The clear next step is to leverage reinforcement learning (RL) to enable the agent to autonomously discover optimal and potentially non-obvious folding policies by directly optimizing for task success.

\clearpage
\bibliography{biblio}
\bibliographystyle{colm2024_conference}

\newpage

\appendix
\section{Appendix}

\subsection{Case Study}
\label{app:case_study}

\begin{figure}[t]
    \centering
    \includegraphics[width=1.0\linewidth]{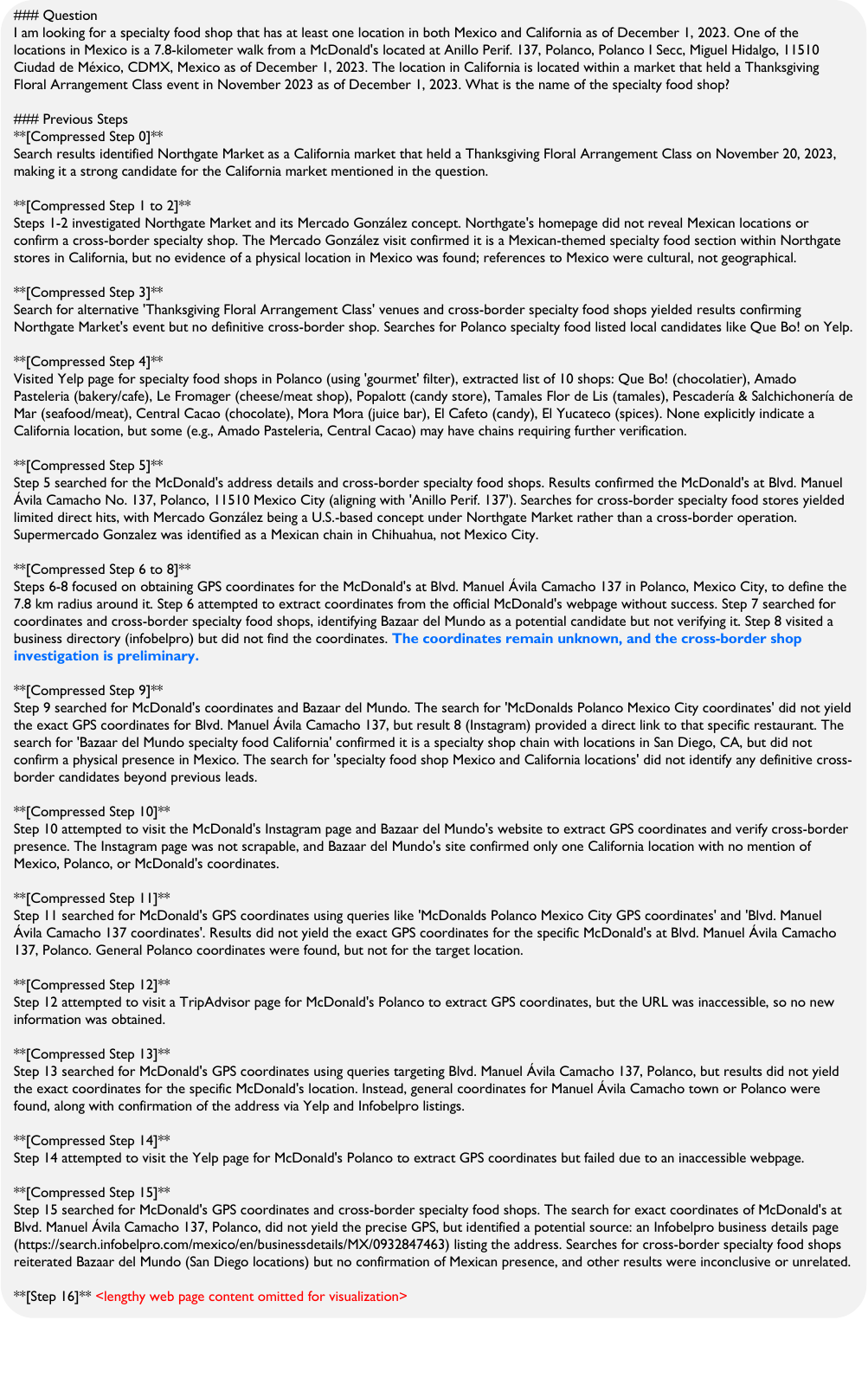}
    \vspace{-16mm}
    \caption{Context of case 1 at step 17.}
    \label{fig:case1_context}
\end{figure}

\begin{figure}[t]
    \centering
    \includegraphics[width=1.0\linewidth]{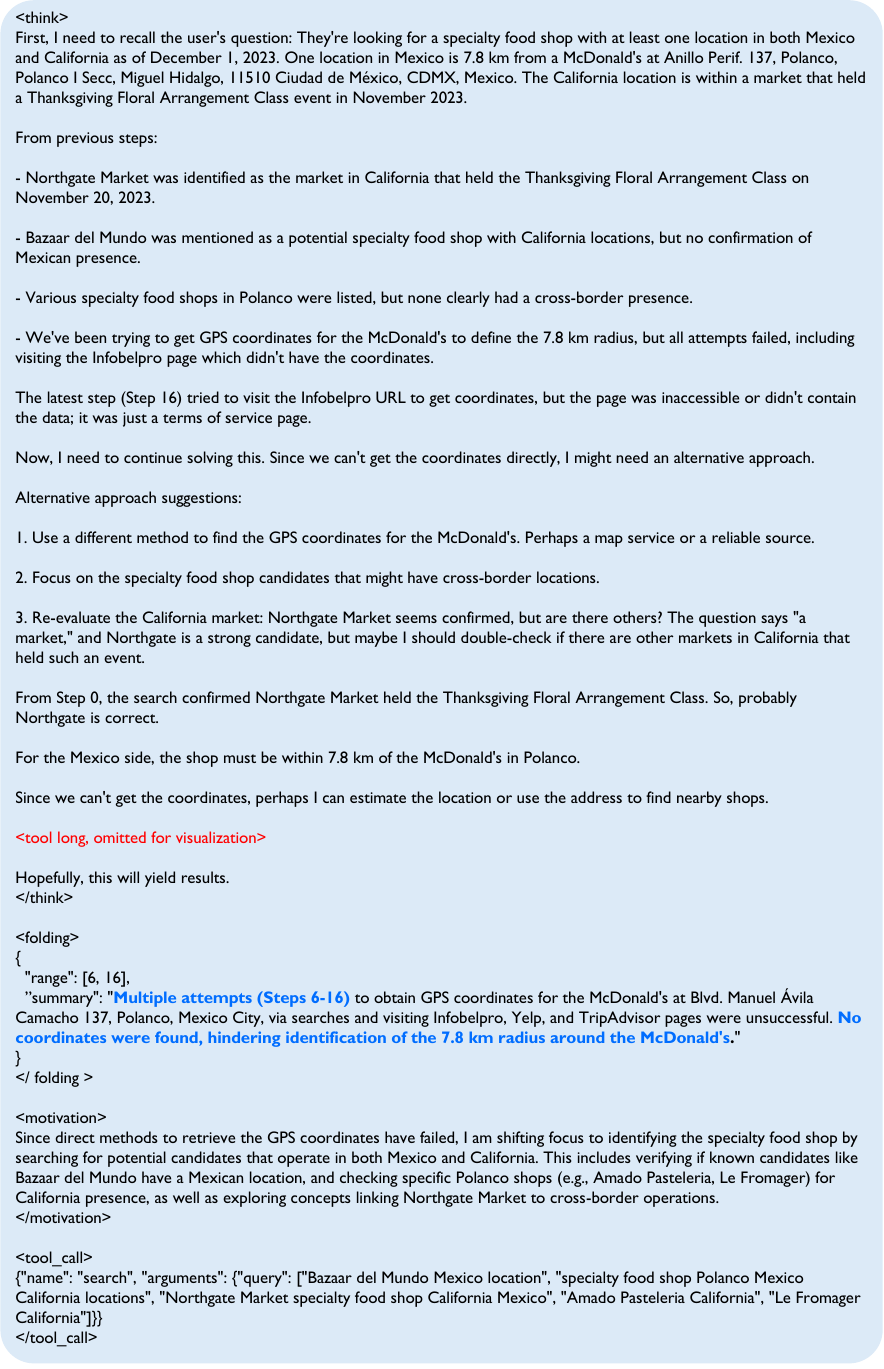}
    \vspace{-9mm}
    \caption{Response of case 1 at step 17.}
    \label{fig:case1_response}
\end{figure}

\begin{figure}[t]
    \centering
    \includegraphics[width=1.0\linewidth]{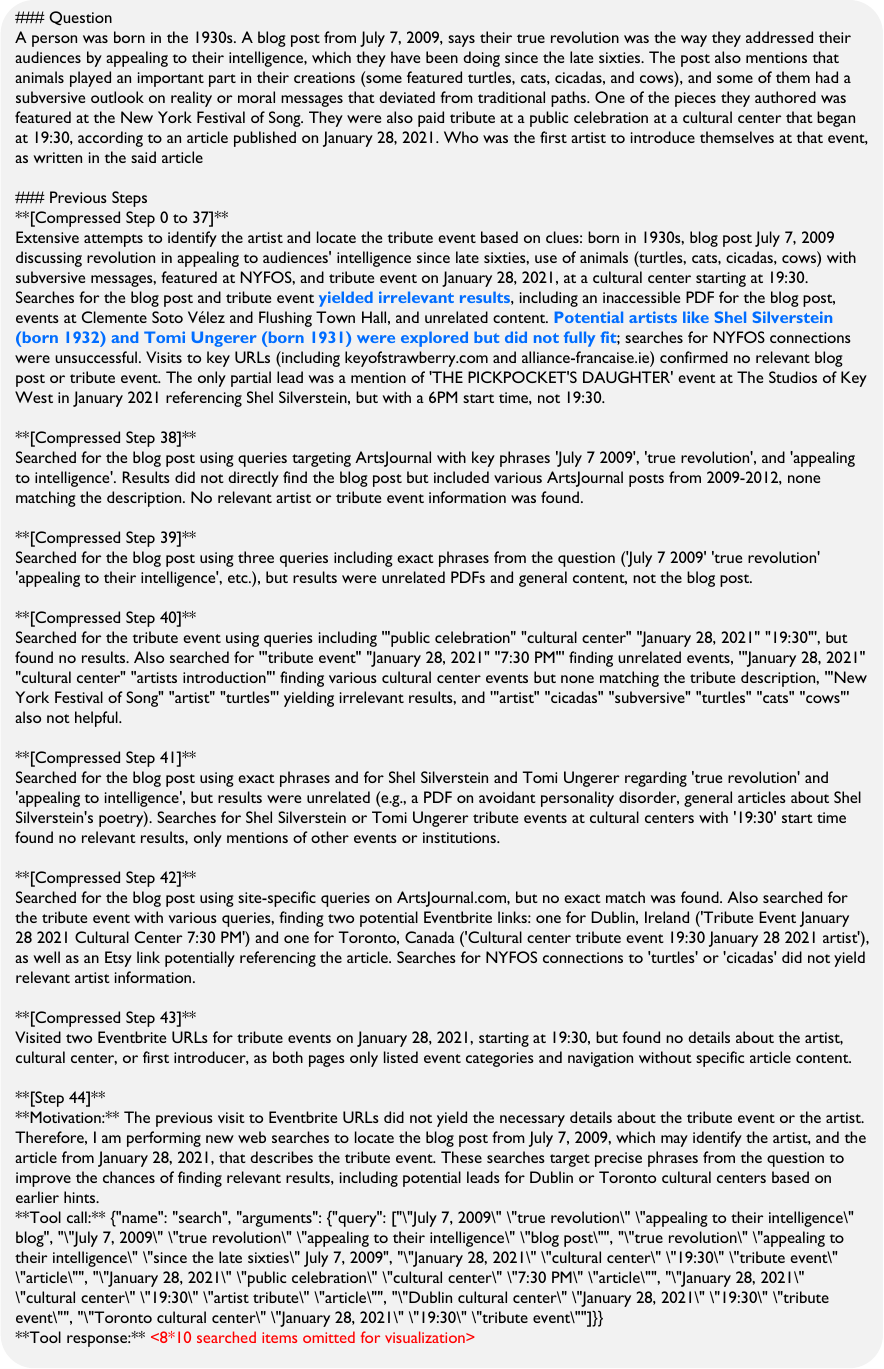}
    \vspace{-8mm}
    \caption{Context of case 2 at step 45.}
    \label{fig:case2_context}
\end{figure}

\begin{figure}[t]
    \centering
    \includegraphics[width=1.0\linewidth]{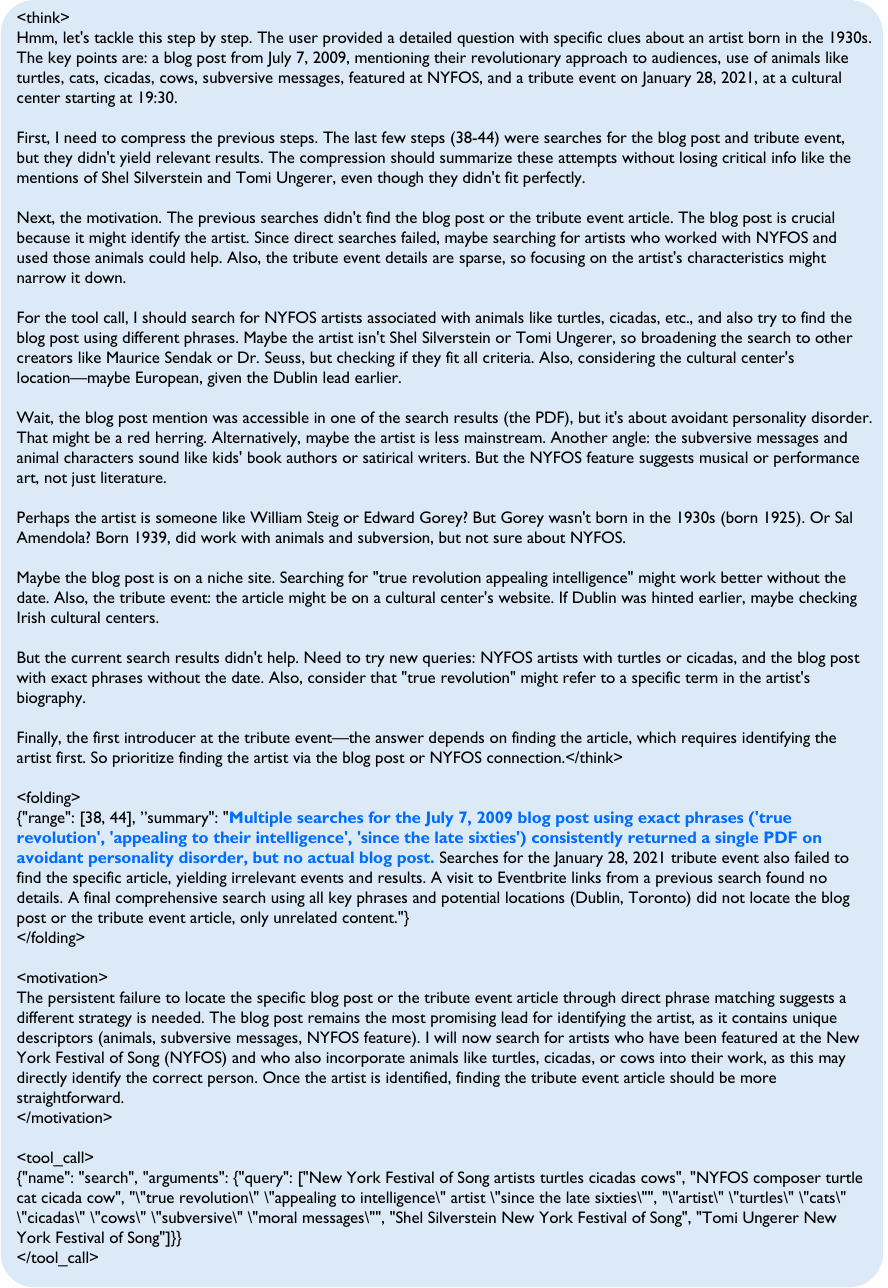}
    \vspace{-4mm}
    \caption{Response of case 2 at step 45.}
    \label{fig:case2_response}
\end{figure}

We provide two cases here.
See case 1 in Table~\ref{tab:case1_traj}, Figure~\ref{fig:case1_context}, and \ref{fig:case1_response}.
See case 2 in Table~\ref{tab:case2_traj}, Figure~\ref{fig:case2_context}, and \ref{fig:case2_response}.

\begin{longtable}{l|p{11.5cm}}
    \caption{Case Study 1}
    \label{tab:case1_traj} \\

    \toprule
    \textbf{Turn ID} & \textbf{Context} \\
    \midrule 
    \endfirsthead 

    \caption{(Continued) Case Study} \\
    \toprule
    \textbf{Turn ID} & \textbf{Context} \\
    \midrule
    \endhead 

    \bottomrule
    \endlastfoot 

    \multicolumn{2}{r}{\textit{(Continued on next page)}} \\
    \endfoot 
Turn 1 &  \\ \midrule
Turn 2 & [Step 0] \\ \midrule
Turn 3 & [Compressed Step 0], [Step 1] \\ \midrule
Turn 4 & [Compressed Step 0], [Compressed Step 1], [Step 2] \\ \midrule
Turn 5 & [Compressed Step 0], [Compressed Step 1 to 2], [Step 3] \\ \midrule
Turn 6 & [Compressed Step 0], [Compressed Step 1 to 2], [Compressed Step 3], [Step 4] \\ \midrule
Turn 7 & [Compressed Step 0], [Compressed Step 1 to 2], [Compressed Step 3], [Compressed Step 4], [Step 5] \\ \midrule
Turn 8 & [Compressed Step 0], [Compressed Step 1 to 2], [Compressed Step 3], [Compressed Step 4], [Compressed Step 5], [Step 6] \\ \midrule
Turn 9 & [Compressed Step 0], [Compressed Step 1 to 2], [Compressed Step 3], [Compressed Step 4], [Compressed Step 5], [Compressed Step 6], [Step 7] \\ \midrule
Turn 10 & [Compressed Step 0], [Compressed Step 1 to 2], [Compressed Step 3], [Compressed Step 4], [Compressed Step 5], [Compressed Step 6], [Compressed Step 7], [Step 8] \\ \midrule
Turn 11 & [Compressed Step 0], [Compressed Step 1 to 2], [Compressed Step 3], [Compressed Step 4], [Compressed Step 5], [Compressed Step 6 to 8], [Step 9] \\ \midrule
Turn 12 & [Compressed Step 0], [Compressed Step 1 to 2], [Compressed Step 3], [Compressed Step 4], [Compressed Step 5], [Compressed Step 6 to 8], [Compressed Step 9], [Step 10] \\ \midrule
Turn 13 & [Compressed Step 0], [Compressed Step 1 to 2], [Compressed Step 3], [Compressed Step 4], [Compressed Step 5], [Compressed Step 6 to 8], [Compressed Step 9], [Compressed Step 10], [Step 11] \\ \midrule
Turn 14 & [Compressed Step 0], [Compressed Step 1 to 2], [Compressed Step 3], [Compressed Step 4], [Compressed Step 5], [Compressed Step 6 to 8], [Compressed Step 9], [Compressed Step 10], [Compressed Step 11], [Step 12] \\ \midrule
Turn 15 & [Compressed Step 0], [Compressed Step 1 to 2], [Compressed Step 3], [Compressed Step 4], [Compressed Step 5], [Compressed Step 6 to 8], [Compressed Step 9], [Compressed Step 10], [Compressed Step 11], [Compressed Step 12], [Step 13] \\ \midrule
Turn 16 & [Compressed Step 0], [Compressed Step 1 to 2], [Compressed Step 3], [Compressed Step 4], [Compressed Step 5], [Compressed Step 6 to 8], [Compressed Step 9], [Compressed Step 10], [Compressed Step 11], [Compressed Step 12], [Compressed Step 13], [Step 14] \\ \midrule
Turn 17 & [Compressed Step 0], [Compressed Step 1 to 2], [Compressed Step 3], [Compressed Step 4], [Compressed Step 5], [Compressed Step 6 to 8], [Compressed Step 9], [Compressed Step 10], [Compressed Step 11], [Compressed Step 12], [Compressed Step 13], [Compressed Step 14], [Step 15] \\ \midrule
Turn 18 & [Compressed Step 0], [Compressed Step 1 to 2], [Compressed Step 3], [Compressed Step 4], [Compressed Step 5], [Compressed Step 6 to 8], [Compressed Step 9], [Compressed Step 10], [Compressed Step 11], [Compressed Step 12], [Compressed Step 13], [Compressed Step 14], [Compressed Step 15], [Step 16] \\ \midrule
Turn 19 & [Compressed Step 0], [Compressed Step 1 to 2], [Compressed Step 3], [Compressed Step 4], [Compressed Step 5], [Compressed Step 6 to 16], [Step 17] \\ \midrule
Turn 20 & [Compressed Step 0], [Compressed Step 1 to 2], [Compressed Step 3], [Compressed Step 4], [Compressed Step 5], [Compressed Step 6 to 16], [Compressed Step 17], [Step 18] \\ \midrule
Turn 21 & [Compressed Step 0], [Compressed Step 1 to 2], [Compressed Step 3], [Compressed Step 4], [Compressed Step 5], [Compressed Step 6 to 16], [Compressed Step 17], [Compressed Step 18], [Step 19] \\ \midrule
Turn 22 & [Compressed Step 0], [Compressed Step 1 to 2], [Compressed Step 3], [Compressed Step 4], [Compressed Step 5], [Compressed Step 6 to 16], [Compressed Step 17], [Compressed Step 18], [Compressed Step 19], [Step 20] \\ \midrule
Turn 23 & [Compressed Step 0], [Compressed Step 1 to 2], [Compressed Step 3], [Compressed Step 4], [Compressed Step 5], [Compressed Step 6 to 16], [Compressed Step 17], [Compressed Step 18], [Compressed Step 19], [Compressed Step 20], [Step 21] \\ \midrule
Turn 24 & [Compressed Step 0], [Compressed Step 1 to 2], [Compressed Step 3], [Compressed Step 4], [Compressed Step 5], [Compressed Step 6 to 16], [Compressed Step 17], [Compressed Step 18], [Compressed Step 19], [Compressed Step 20], [Compressed Step 21], [Step 22] \\ \midrule
Turn 25 & [Compressed Step 0], [Compressed Step 1 to 2], [Compressed Step 3], [Compressed Step 4], [Compressed Step 5], [Compressed Step 6 to 16], [Compressed Step 17], [Compressed Step 18], [Compressed Step 19], [Compressed Step 20], [Compressed Step 21], [Compressed Step 22], [Step 23] \\ \midrule
Turn 26 & [Compressed Step 0], [Compressed Step 1 to 2], [Compressed Step 3], [Compressed Step 4], [Compressed Step 5], [Compressed Step 6 to 16], [Compressed Step 17], [Compressed Step 18], [Compressed Step 19], [Compressed Step 20], [Compressed Step 21], [Compressed Step 22], [Compressed Step 23], [Step 24] \\ \midrule
Turn 27 & [Compressed Step 0], [Compressed Step 1 to 2], [Compressed Step 3], [Compressed Step 4], [Compressed Step 5], [Compressed Step 6 to 16], [Compressed Step 17], [Compressed Step 18], [Compressed Step 19], [Compressed Step 20], [Compressed Step 21], [Compressed Step 22], [Compressed Step 23], [Compressed Step 24], [Step 25] \\ \midrule
Turn 28 & [Compressed Step 0], [Compressed Step 1 to 2], [Compressed Step 3], [Compressed Step 4], [Compressed Step 5], [Compressed Step 6 to 16], [Compressed Step 17], [Compressed Step 18], [Compressed Step 19], [Compressed Step 20], [Compressed Step 21], [Compressed Step 22], [Compressed Step 23], [Compressed Step 24], [Compressed Step 25], [Step 26] \\ \midrule
Turn 29 & [Compressed Step 0], [Compressed Step 1 to 2], [Compressed Step 3], [Compressed Step 4], [Compressed Step 5], [Compressed Step 6 to 16], [Compressed Step 17], [Compressed Step 18], [Compressed Step 19], [Compressed Step 20], [Compressed Step 21], [Compressed Step 22], [Compressed Step 23], [Compressed Step 24], [Compressed Step 25], [Compressed Step 26], [Step 27] \\ \midrule
Turn 30 & [Compressed Step 0], [Compressed Step 1 to 2], [Compressed Step 3], [Compressed Step 4], [Compressed Step 5], [Compressed Step 6 to 16], [Compressed Step 17], [Compressed Step 18], [Compressed Step 19], [Compressed Step 20], [Compressed Step 21 to 27], [Step 28] \\ \midrule
Turn 31 & [Compressed Step 0], [Compressed Step 1 to 2], [Compressed Step 3], [Compressed Step 4], [Compressed Step 5], [Compressed Step 6 to 16], [Compressed Step 17], [Compressed Step 18], [Compressed Step 19], [Compressed Step 20], [Compressed Step 21 to 27], [Compressed Step 28], [Step 29] \\ \midrule
Turn 32 & [Compressed Step 0], [Compressed Step 1 to 2], [Compressed Step 3], [Compressed Step 4], [Compressed Step 5], [Compressed Step 6 to 16], [Compressed Step 17], [Compressed Step 18], [Compressed Step 19], [Compressed Step 20], [Compressed Step 21 to 27], [Compressed Step 28], [Compressed Step 29], [Step 30] \\ \midrule
Turn 33 & [Compressed Step 0], [Compressed Step 1 to 2], [Compressed Step 3], [Compressed Step 4], [Compressed Step 5], [Compressed Step 6 to 16], [Compressed Step 17], [Compressed Step 18], [Compressed Step 19], [Compressed Step 20], [Compressed Step 21 to 27], [Compressed Step 28], [Compressed Step 29], [Compressed Step 30], [Step 31] \\ \midrule
Turn 34 & [Compressed Step 0], [Compressed Step 1 to 2], [Compressed Step 3], [Compressed Step 4], [Compressed Step 5], [Compressed Step 6 to 16], [Compressed Step 17], [Compressed Step 18], [Compressed Step 19], [Compressed Step 20], [Compressed Step 21 to 27], [Compressed Step 28], [Compressed Step 29], [Compressed Step 30], [Compressed Step 31], [Step 32] \\ \midrule
Turn 35 & [Compressed Step 0], [Compressed Step 1 to 2], [Compressed Step 3], [Compressed Step 4], [Compressed Step 5], [Compressed Step 6 to 16], [Compressed Step 17], [Compressed Step 18], [Compressed Step 19], [Compressed Step 20], [Compressed Step 21 to 27], [Compressed Step 28], [Compressed Step 29], [Compressed Step 30], [Compressed Step 31], [Compressed Step 32], [Step 33] \\ 
\end{longtable}

\begin{longtable}{l|p{11.5cm}}
    \caption{Case Study 2}
    \label{tab:case2_traj} \\

    \toprule
    \textbf{Turn ID} & \textbf{Context} \\
    \midrule 
    \endfirsthead 

    \caption{(Continued) Case Study 2} \\
    \toprule
    \textbf{Turn ID} & \textbf{Context} \\
    \midrule
    \endhead 

    \bottomrule
    \endlastfoot 

    \multicolumn{2}{r}{\textit{(Continued on next page)}} \\
    \endfoot 
Turn 1 &  \\ \midrule
Turn 2 & [Step 0] \\ \midrule
Turn 3 & [Compressed Step 0], [Step 1] \\ \midrule
Turn 4 & [Compressed Step 0 to 1], [Step 2] \\ \midrule
Turn 5 & [Compressed Step 0 to 2], [Step 3] \\ \midrule
Turn 6 & [Compressed Step 0 to 2], [Compressed Step 3], [Step 4] \\ \midrule
Turn 7 & [Compressed Step 0 to 2], [Compressed Step 3], [Compressed Step 4], [Step 5] \\ \midrule
Turn 8 & [Compressed Step 0 to 2], [Compressed Step 3], [Compressed Step 4], [Compressed Step 5], [Step 6] \\ \midrule
Turn 9 & [Compressed Step 0 to 2], [Compressed Step 3], [Compressed Step 4], [Compressed Step 5], [Compressed Step 6], [Step 7] \\ \midrule
Turn 10 & [Compressed Step 0 to 2], [Compressed Step 3], [Compressed Step 4], [Compressed Step 5], [Compressed Step 6], [Compressed Step 7], [Step 8] \\ \midrule
Turn 11 & [Compressed Step 0 to 2], [Compressed Step 3], [Compressed Step 4], [Compressed Step 5], [Compressed Step 6], [Compressed Step 7], [Compressed Step 8], [Step 9] \\ \midrule
Turn 12 & [Compressed Step 0 to 9], [Step 10] \\ \midrule
Turn 13 & [Compressed Step 0 to 9], [Compressed Step 10], [Step 11] \\ \midrule
Turn 14 & [Compressed Step 0 to 9], [Compressed Step 10], [Compressed Step 11], [Step 12] \\ \midrule
Turn 15 & [Compressed Step 0 to 9], [Compressed Step 10], [Compressed Step 11], [Compressed Step 12], [Step 13] \\ \midrule
Turn 16 & [Compressed Step 0 to 9], [Compressed Step 10], [Compressed Step 11], [Compressed Step 12], [Compressed Step 13], [Step 14] \\ \midrule
Turn 17 & [Compressed Step 0 to 9], [Compressed Step 10], [Compressed Step 11], [Compressed Step 12], [Compressed Step 13], [Compressed Step 14], [Step 15] \\ \midrule
Turn 18 & [Compressed Step 0 to 9], [Compressed Step 10], [Compressed Step 11], [Compressed Step 12], [Compressed Step 13], [Compressed Step 14], [Compressed Step 15], [Step 16] \\ \midrule
Turn 19 & [Compressed Step 0 to 9], [Compressed Step 10], [Compressed Step 11], [Compressed Step 12], [Compressed Step 13], [Compressed Step 14], [Compressed Step 15], [Compressed Step 16], [Step 17] \\ \midrule
Turn 20 & [Compressed Step 0 to 9], [Compressed Step 10], [Compressed Step 11], [Compressed Step 12], [Compressed Step 13], [Compressed Step 14], [Compressed Step 15], [Compressed Step 16], [Compressed Step 17], [Step 18] \\ \midrule
Turn 21 & [Compressed Step 0 to 9], [Compressed Step 10], [Compressed Step 11], [Compressed Step 12], [Compressed Step 13], [Compressed Step 14], [Compressed Step 15], [Compressed Step 16], [Compressed Step 17], [Compressed Step 18], [Step 19] \\ \midrule
Turn 22 & [Compressed Step 0 to 9], [Compressed Step 10], [Compressed Step 11], [Compressed Step 12], [Compressed Step 13], [Compressed Step 14], [Compressed Step 15], [Compressed Step 16], [Compressed Step 17], [Compressed Step 18], [Compressed Step 19], [Step 20] \\ \midrule
Turn 23 & [Compressed Step 0 to 9], [Compressed Step 10], [Compressed Step 11], [Compressed Step 12], [Compressed Step 13], [Compressed Step 14], [Compressed Step 15], [Compressed Step 16], [Compressed Step 17], [Compressed Step 18], [Compressed Step 19], [Compressed Step 20], [Step 21] \\ \midrule
Turn 24 & [Compressed Step 0 to 9], [Compressed Step 10], [Compressed Step 11], [Compressed Step 12], [Compressed Step 13], [Compressed Step 14], [Compressed Step 15], [Compressed Step 16], [Compressed Step 17], [Compressed Step 18], [Compressed Step 19], [Compressed Step 20], [Compressed Step 21], [Step 22] \\ \midrule
Turn 25 & [Compressed Step 0 to 22], [Step 23] \\ \midrule
Turn 26 & [Compressed Step 0 to 22], [Compressed Step 23], [Step 24] \\ \midrule
Turn 27 & [Compressed Step 0 to 22], [Compressed Step 23], [Compressed Step 24], [Step 25] \\ \midrule
Turn 28 & [Compressed Step 0 to 22], [Compressed Step 23], [Compressed Step 24], [Compressed Step 25], [Step 26] \\ \midrule
Turn 29 & [Compressed Step 0 to 22], [Compressed Step 23], [Compressed Step 24], [Compressed Step 25], [Compressed Step 26], [Step 27] \\ \midrule
Turn 30 & [Compressed Step 0 to 22], [Compressed Step 23], [Compressed Step 24], [Compressed Step 25], [Compressed Step 26], [Compressed Step 27], [Step 28] \\ \midrule
Turn 31 & [Compressed Step 0 to 22], [Compressed Step 23], [Compressed Step 24], [Compressed Step 25], [Compressed Step 26], [Compressed Step 27], [Compressed Step 28], [Step 29] \\ \midrule
Turn 32 & [Compressed Step 0 to 22], [Compressed Step 23], [Compressed Step 24], [Compressed Step 25], [Compressed Step 26], [Compressed Step 27], [Compressed Step 28], [Compressed Step 29], [Step 30] \\ \midrule
Turn 33 & [Compressed Step 0 to 22], [Compressed Step 23], [Compressed Step 24], [Compressed Step 25], [Compressed Step 26], [Compressed Step 27], [Compressed Step 28], [Compressed Step 29], [Compressed Step 30], [Step 31] \\ \midrule
Turn 34 & [Compressed Step 0 to 22], [Compressed Step 23], [Compressed Step 24], [Compressed Step 25], [Compressed Step 26], [Compressed Step 27], [Compressed Step 28], [Compressed Step 29], [Compressed Step 30], [Compressed Step 31], [Step 32] \\ \midrule
Turn 35 & [Compressed Step 0 to 22], [Compressed Step 23], [Compressed Step 24], [Compressed Step 25], [Compressed Step 26], [Compressed Step 27], [Compressed Step 28], [Compressed Step 29], [Compressed Step 30], [Compressed Step 31], [Compressed Step 32], [Step 33] \\ \midrule
Turn 36 & [Compressed Step 0 to 22], [Compressed Step 23], [Compressed Step 24], [Compressed Step 25], [Compressed Step 26], [Compressed Step 27], [Compressed Step 28], [Compressed Step 29], [Compressed Step 30], [Compressed Step 31], [Compressed Step 32], [Compressed Step 33], [Step 34] \\ \midrule
Turn 37 & [Compressed Step 0 to 34], [Step 35] \\ \midrule
Turn 38 & [Compressed Step 0 to 34], [Compressed Step 35], [Step 36] \\ \midrule
Turn 39 & [Compressed Step 0 to 34], [Compressed Step 35], [Compressed Step 36], [Step 37] \\ \midrule
Turn 40 & [Compressed Step 0 to 34], [Compressed Step 35], [Compressed Step 36], [Compressed Step 37], [Step 38] \\ \midrule
Turn 41 & [Compressed Step 0 to 34], [Compressed Step 35], [Compressed Step 36], [Compressed Step 37], [Compressed Step 38], [Step 39] \\ \midrule
Turn 42 & [Compressed Step 0 to 34], [Compressed Step 35], [Compressed Step 36], [Compressed Step 37], [Compressed Step 38], [Compressed Step 39], [Step 40] \\ \midrule
Turn 43 & [Compressed Step 0 to 34], [Compressed Step 35], [Compressed Step 36], [Compressed Step 37], [Compressed Step 38], [Compressed Step 39], [Compressed Step 40], [Step 41] \\ \midrule
Turn 44 & [Compressed Step 0 to 34], [Compressed Step 35], [Compressed Step 36], [Compressed Step 37], [Compressed Step 38], [Compressed Step 39], [Compressed Step 40], [Compressed Step 41], [Step 42] \\ \midrule
Turn 45 & [Compressed Step 0 to 34], [Compressed Step 35], [Compressed Step 36], [Compressed Step 37], [Compressed Step 38], [Compressed Step 39], [Compressed Step 40], [Compressed Step 41], [Compressed Step 42], [Step 43] \\ \midrule
Turn 46 & [Compressed Step 0 to 34], [Compressed Step 35], [Compressed Step 36], [Compressed Step 37], [Compressed Step 38], [Compressed Step 39], [Compressed Step 40], [Compressed Step 41], [Compressed Step 42], [Compressed Step 43], [Step 44] \\ \midrule
Turn 47 & [Compressed Step 0 to 44], [Step 45] \\ \midrule
Turn 48 & [Compressed Step 0 to 44], [Compressed Step 45], [Step 46] \\ \midrule
Turn 49 & [Compressed Step 0 to 44], [Compressed Step 45], [Compressed Step 46], [Step 47] \\ \midrule
Turn 50 & [Compressed Step 0 to 44], [Compressed Step 45], [Compressed Step 46], [Compressed Step 47], [Step 48] \\ \midrule
Turn 51 & [Compressed Step 0 to 44], [Compressed Step 45], [Compressed Step 46], [Compressed Step 47], [Compressed Step 48], [Step 49] \\ \midrule
Turn 52 & [Compressed Step 0 to 44], [Compressed Step 45], [Compressed Step 46], [Compressed Step 47], [Compressed Step 48], [Compressed Step 49], [Step 50] \\ \midrule
Turn 53 & [Compressed Step 0 to 44], [Compressed Step 45], [Compressed Step 46], [Compressed Step 47], [Compressed Step 48], [Compressed Step 49], [Compressed Step 50], [Step 51] \\ \midrule
Turn 54 & [Compressed Step 0 to 44], [Compressed Step 45], [Compressed Step 46], [Compressed Step 47], [Compressed Step 48], [Compressed Step 49], [Compressed Step 50], [Compressed Step 51], [Step 52] \\ \midrule
Turn 55 & [Compressed Step 0 to 44], [Compressed Step 45], [Compressed Step 46], [Compressed Step 47], [Compressed Step 48], [Compressed Step 49], [Compressed Step 50], [Compressed Step 51], [Compressed Step 52], [Step 53] \\ \midrule
Turn 56 & [Compressed Step 0 to 44], [Compressed Step 45], [Compressed Step 46], [Compressed Step 47], [Compressed Step 48], [Compressed Step 49], [Compressed Step 50], [Compressed Step 51], [Compressed Step 52], [Compressed Step 53], [Step 54] \\ \midrule
Turn 57 & [Compressed Step 0 to 44], [Compressed Step 45], [Compressed Step 46], [Compressed Step 47], [Compressed Step 48], [Compressed Step 49], [Compressed Step 50], [Compressed Step 51], [Compressed Step 52], [Compressed Step 53], [Compressed Step 54], [Step 55] \\ \midrule
Turn 58 & [Compressed Step 0 to 44], [Compressed Step 45], [Compressed Step 46], [Compressed Step 47], [Compressed Step 48], [Compressed Step 49], [Compressed Step 50], [Compressed Step 51], [Compressed Step 52], [Compressed Step 53], [Compressed Step 54], [Compressed Step 55], [Step 56] \\ \midrule
Turn 59 & [Compressed Step 0 to 44], [Compressed Step 45], [Compressed Step 46], [Compressed Step 47], [Compressed Step 48], [Compressed Step 49], [Compressed Step 50], [Compressed Step 51], [Compressed Step 52], [Compressed Step 53], [Compressed Step 54], [Compressed Step 55], [Compressed Step 56], [Step 57] \\ \midrule
Turn 60 & [Compressed Step 0 to 44], [Compressed Step 45], [Compressed Step 46], [Compressed Step 47], [Compressed Step 48], [Compressed Step 49], [Compressed Step 50], [Compressed Step 51], [Compressed Step 52], [Compressed Step 53], [Compressed Step 54], [Compressed Step 55], [Compressed Step 56], [Compressed Step 57], [Step 58] \\
\end{longtable}

\end{document}